\documentclass[10pt,twocolumn,letterpaper]{article}

\usepackage[pagenumbers]{cvpr} 
\usepackage{arydshln}
\definecolor{cvprblue}{rgb}{0.21,0.49,0.74}
\usepackage[pagebackref,breaklinks,colorlinks,allcolors=cvprblue]{hyperref}
\usepackage{balance}
\def\paperID{*****} 
\def\confName{CVPR}
\def\confYear{2025}

\title{Technical Report for Egocentric Mistake Detection for the HoloAssist Challenge}

\author{Constantin Patsch, Marsil Zakour, Yuankai Wu, Eckehard Steinbach\\
Technical University of Munich\\
{\tt\small \{constantin.patsch, marsil.zakour,yuankai.wu,eckehard.steinbach\}@tum.de}
}

\begin{document}
\maketitle
\begin{abstract}
In this report, we address the task of online mistake detection, which is vital in domains like industrial automation and education, where real-time video analysis allows human operators to correct errors as they occur. While previous work focuses on procedural errors involving action order, broader error types must be addressed for real-world use. We introduce an online mistake detection framework that handles both procedural and execution errors (e.g., motor slips or tool misuse). Upon detecting an error, we use a large language model (LLM) to generate explanatory feedback. Experiments on the HoloAssist benchmark confirm the effectiveness of our approach, where our approach is placed second on the mistake detection task.
\end{abstract}    
\vspace{-8pt}
\section{Introduction}
\label{sec:intro}
Action detection and recognition methods~\cite{wang2021oadtr, phan2024zeetad, patsch2023self, yang2022colar} accurately interpret human actions from video by leveraging spatiotemporal cues. However, a truly intelligent assistant should go beyond recognition and assess the correctness of actions to help users perform tasks accurately. Such systems hold potential for both everyday scenarios like cooking or home maintenance, and industrial applications, such as assembly or mechanical repair.

Following recent online mistake detection efforts~\cite{lee2024error, seminara2024differentiable, flaborea2024prego}, systems should support real-time inference to identify errors from a continuous video stream, enabling users to react promptly and avoid further consequences. An egocentric perspective is particularly beneficial, capturing task execution from the user's viewpoint and avoiding occlusions common in static views.

Recent methods largely focus on procedural errors—like incorrect sequencing or missed actions~\cite{lee2024error, seminara2024differentiable, flaborea2024prego}—but real-world tasks often involve a broader range of errors. Partial action ordering alone may not fully reflect task success. By also considering execution errors, such as motor mistakes or misuse of tools, a more complete view of task performance emerges. 
Most current methods further only detect mistakes~\cite{flaborea2024prego, seminara2024differentiable} without providing insights. However, understanding why an action is wrong, such as missing prerequisites or incorrect execution, can help users correct their behavior. To this end, we leverage recent advances in language and vision-language models~\cite{touvron2023llama, Maaz2023VideoChatGPT, li2023blip} to generate explanations for detected mistakes using an LLM.

Thus, within our approach, we focus on capturing procedural errors that mainly relate to the relative ordering of actions and depend on temporal relations, as well as execution errors, which refer to how the human is performing certain actions.
As a result, our approach can better capture the correct execution of a task while being versatile concerning varying error types.
Our contribution is three-fold:
\begin{itemize}
\item We introduce an online mistake-detection method that captures both procedural and execution errors, enabling a more holistic assessment of task execution beyond temporal ordering alone.
\item Our approach generates explanations for detected mistakes, offering users actionable insights and facilitating error resolution.
\item Our approach achieves the second place on the Holo-Assist~\cite{Wang_2023_ICCV} benchmark\footnote{\url{https://www.codabench.org/competitions/2613/}}.
\end{itemize}

\section{Methodology}
In this section, we give an overview of the design of our approach and explain the individual components with respect to the mistake detection and error explanation tasks. 

\begin{figure*}
    \centering
    \includegraphics[width=0.93\linewidth]{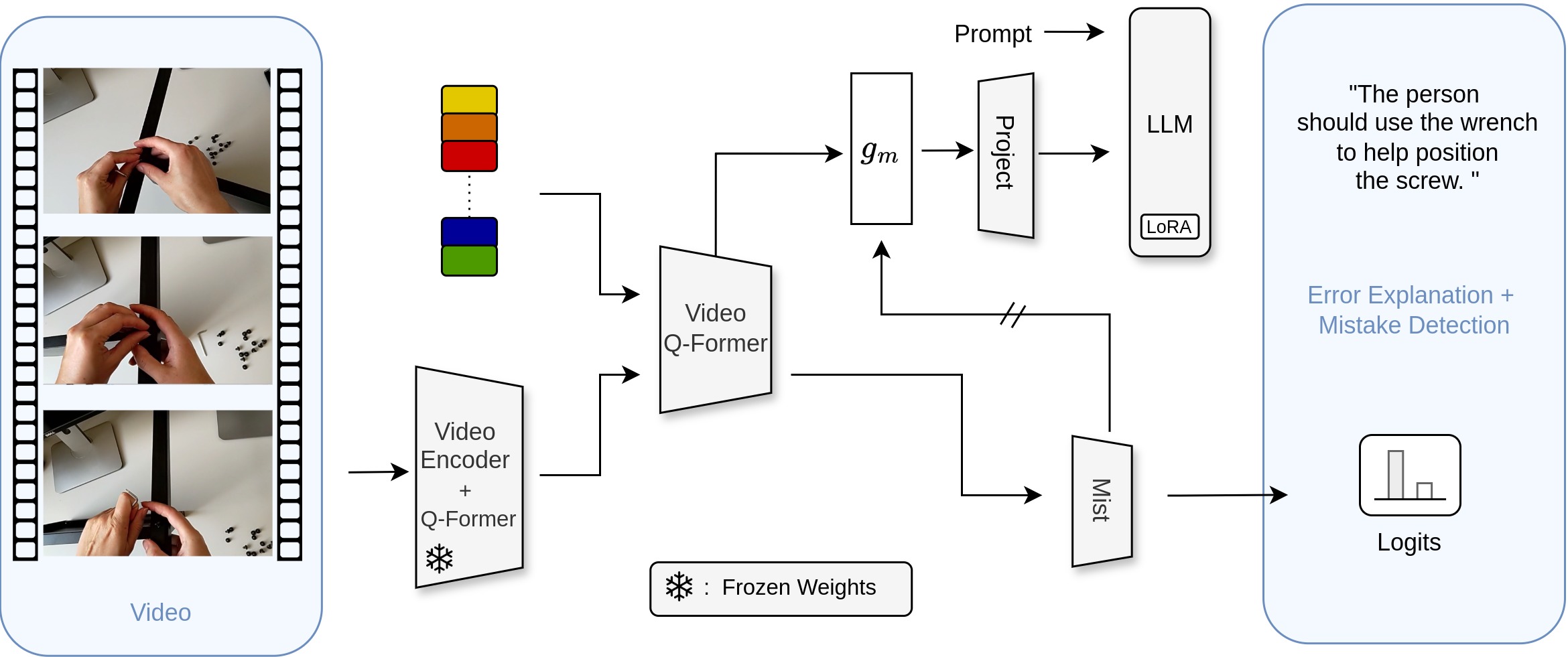}
    \caption{Overview: Our method processes a continuous RGB stream using a video encoder and a Q-Former to extract framewise visual relations. A Video Q-Former captures temporal dependencies between the features, using learnable queries (colored squares) inspired by~\cite{li2023blip}. These features are fed to the mistake classification layer for final predictions. The Video Q-Former output is gated via $g_m$ and projected into the LLM embedding space to generate error explanations. Double slashes indicate layers without backpropagation.}
    \label{fig:approach}
\end{figure*}

\subsection{Mistake Detection}
As illustrated in \Cref{fig:approach}, our online setup processes a continuous video stream by dividing the input into frame segments up to the current timestep $t$. These frames are fed into a visual encoder in combination with a Q-Former, which is based on the architecture introduced in Blip2~\cite{li2023blip}, to extract visual features. In our implementation, the video encoder is a Vision Transformer (ViT)~\cite{dosovitskiy2021an}. We represent the resulting segment of visual features as $v=[v_{t-t_s}, ..., v_{t}]$, where $v \in \mathbb{R}^{t_s \times d_1}$. Here, $t_s$ refers to the number of timesteps within the segment, and $d_1$ denotes the dimensionality of the extracted features.

Together with learnable queries $q \in \mathbb{R}^{t_q \times d_2}$, the feature sequence $v$ is fed into the Video Q-Former. While the standard Q-Former focuses on capturing spatial information within individual frames, the Video Q-Former is designed to model temporal dependencies across multiple frame-wise features. Despite this extension, the Q-Former retains the overall architecture proposed in~\cite{li2023blip}, similar to the adaptation shown in~\cite{zhang2023video}, and is likewise built upon a BERT encoder~\cite{devlin2019bert}. The output of this process is a set of temporally-aware features denoted as $f \in \mathbb{R}^{t_q \times d_2}$.


The features $f$ are passed to the mistake classification layer to obtain the final mistake logits $m\in \mathbb{R}^{t_q}$. The error explanation generation process is initiated once a mistake is identified based on those logits.

\subsection{Error Explanation}
The error explanation component aims to offer textual reasoning that clarifies why a particular action is incorrect. These explanations address both execution and procedural mistakes detected within the analyzed segment.

The Video Q-Former features $f$, extracted during the mistake detection phase, are passed through a gating mechanism $g_m$ to generate these explanations. This is formally defined as:

\begin{equation}
g_m(\sigma(m)) = 
\begin{cases} 
1, & \text{if } \sigma(m) \geq \tau \\
0, & \text{otherwise}
\end{cases}
\end{equation}
where $\sigma$ denotes the sigmoid function. When the predicted logit exceeds a predefined threshold $\tau$, the features from the Video Q-Former are passed to a projection layer. This projection layer is a linear transformation that maps the features into the LLM embedding space. The projected features, along with a prompt, are then input to the LLM to generate the final explanation. This mechanism ensures that explanations are produced only when an error is detected.

\begin{table*}
\vskip 2mm
\begin{center}
\begin{tabular}{llccccc}
\toprule
 & \multicolumn{6}{c}{\textbf{Holo Assist}} \\
\midrule
\textbf{Methods} & \textbf{Modality} & \textbf{F1-Score} & \multicolumn{2}{c}{\textbf{Correct}} & \multicolumn{2}{c}{\textbf{Mistake}} \\
 & & & \textbf{Prec} & \textbf{Rec} & \textbf{Prec} & \textbf{Rec} \\

\midrule

{TSformer (Baseline)} & RGB & {35.1} & {82.6} & {51.8} & \textbf{12.9} & \underline{26.9} \\

{TSformer (Baseline)} & RGB + H(GT) & {36.2} & {85.5} & {43.1} & 9.7 & {11.5} \\



GazeCompl (CVPR24) & RGB + E & 51.0 & 95.0 & \textbf{92.0} & {6.0} & {9.0} \\

266097 (CVPR25) & RGB  & 54.0 & \underline{96.0} & 86.0 & 9.0 & 26.0 \\

MR-CAS (CVPR25) & RGB  & \textbf{57.0} & \textbf{97.0} & 60.0 & 8.0 & \textbf{63.0} \\

\hdashline

Ours & RGB & \underline{55.0} & \underline{96.0} & \underline{91.0} & \underline{11.0} & 21.0 \\

\bottomrule
\end{tabular}
\end{center}
\caption{Performance comparison on Holo Assist~\cite{Wang_2023_ICCV}. TSformer~\cite{bertasius2021space} + H(GT)\cite{Wang_2023_ICCV} denotes the TimeSformer model combined with ground truth hand pose as reported by ~\cite{Wang_2023_ICCV}. E denotes the eye gaze information.}
\label{tab:holo_assist}
\end{table*}

\begin{figure*}
    \centering
    \includegraphics[width=0.9\linewidth]{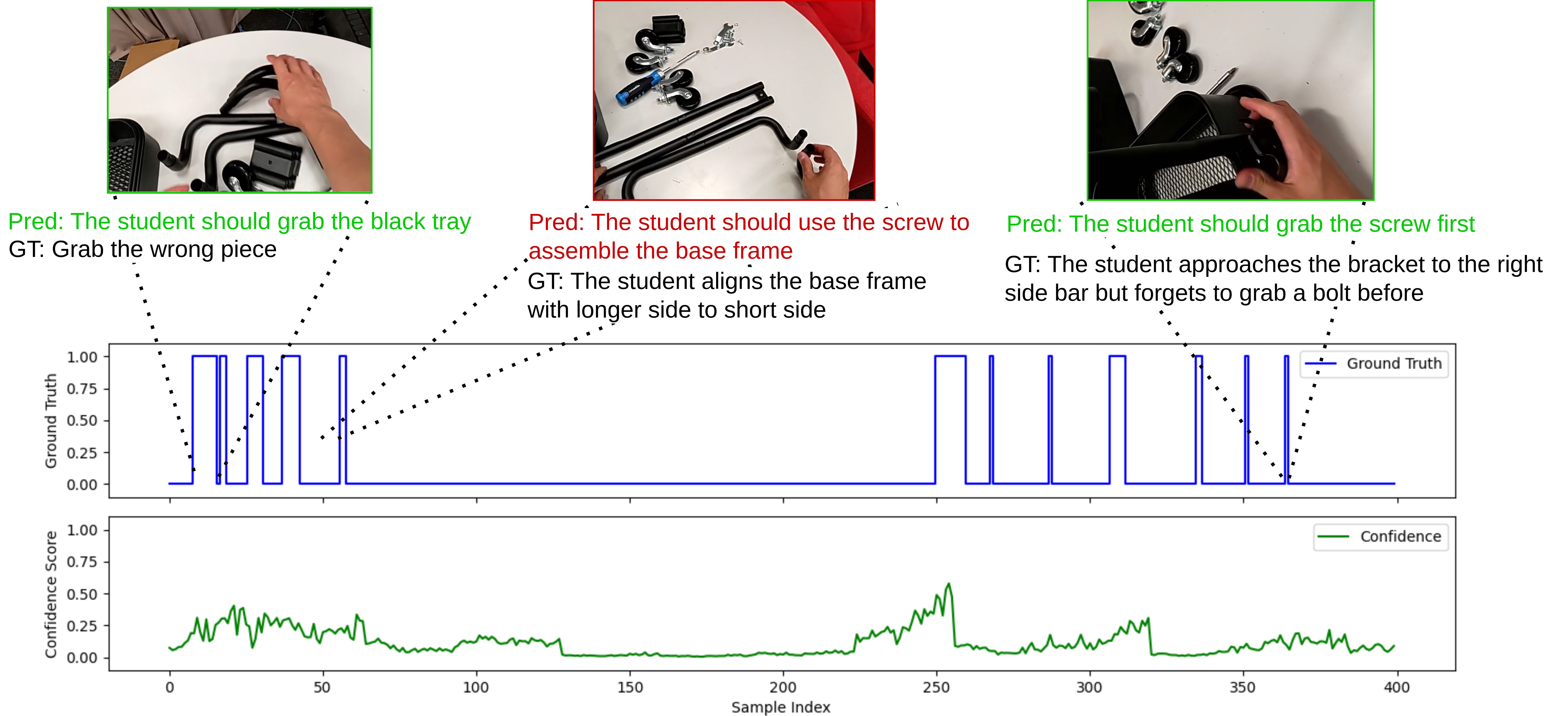}
    \caption{Qualitative results on a video cutout of the HoloAssist dataset, where exemplary frames are sampled from the indicated segments. The confidence scores indicating the mistake detection probabilities $\hat{y}$ and the ground truth indicating the mistake annotations are visualized over time.}
    \label{fig:qualitative}
\end{figure*}

\begin{table}
\vskip 2mm
\begin{center}
\begin{tabular}{lccccc}
\toprule
 \multicolumn{4}{c}{\textbf{Holo Assist}} \\
\midrule
\textbf{Methods} & \multicolumn{1}{c}{\textbf{BLEU}} & \textbf{ROUGEL} & \textbf{CIDEr} \\
\midrule
GT Sampling  & 0.46 &  0.50 &	0.30	\\
Videollama~\cite{zhang2023video} & 0.52 & \underline{0.57} & \underline{0.68} \\
Video-ChatGPT~\cite{Maaz2023VideoChatGPT} & \textbf{0.55} &  \textbf{0.58} &	0.61	\\

Ours & \underline{0.53} & \textbf{0.58} & \textbf{0.76} \\

\bottomrule
\end{tabular}
\end{center}
\caption{Performance comparison on the error explanation generation task for the Holo Assist~\cite{Wang_2023_ICCV} dataset.}\label{tab:explanation_holo}
\end{table}

\section{Experiments}
We provide qualitative and quantitative results of the mistake detection and explanation generation tasks.
\subsection{Mistake Detection}

For the Holo Assist dataset, we report results via the official competition server, as test annotations are unavailable. As shown in \Cref{tab:holo_assist}, our approach significantly outperforms the RGB-only TSformer\cite{bertasius2021space} and improves over GazeCompl~\cite{mazzamuto2024eyes} by $3.8\%$ without relying on eye gaze input. The dataset’s frequent background segments and mix of procedural and execution errors highlight the robustness of our approach under challenging conditions.


\subsection{Error Explanation}
We further evaluate the explanation capabilities of our approach on the Holo-Assist~\cite{Wang_2023_ICCV} dataset, which includes detailed error descriptions like "The battery is upside down" or "The person accidentally turned on the GoPro." As shown in ~\Cref{tab:explanation_holo}, our method matches or surpasses baseline performance, with notable improvements in semantic similarity, particularly reflected by higher CIDEr scores.
\Cref{fig:qualitative} shows a qualitative example of HoloAssist.

\section{Conclusion}
We introduce a versatile online mistake detection approach that also generates mistake explanations. Upon error detection, features are projected into the LLM embedding space to produce explanations. By modeling both spatial execution and temporal action dependencies, our method detects execution and procedural errors.

Experiments on HoloAssist show strong results on both detection and explanation tasks, currently placed second on the benchmark leaderboard.
Future work could address very short mistakes and explore additional modalities to improve detection robustness.

\section{Acknowledgement}
We gratefully acknowledge the funding of the Lighthouse
Initiative Geriatronics by StMWi Bayern (Project X, grant
no. 5140951) and LongLeif GaPa GmbH (Project Y, grant
no. 5140953).

1
1

\balance

{
    \small
    \bibliographystyle{ieeenat_fullname}

}


\end{document}